\documentclass{article} % AfterQuery preprint format (afterquery.sty), arXiv version
\ifdefined\XeTeXrevision\else\ifdefined\pdfoutput\pdfoutput=1\fi\fi
\usepackage[T1]{fontenc}
\usepackage{afterquery}

\usepackage{amsmath,amsfonts,bm}

\def\eqref#1{equation~\ref{#1}}
\def\1{\bm{1}}

\DeclareMathAlphabet{\mathsfit}{\encodingdefault}{\sfdefault}{m}{sl}
\SetMathAlphabet{\mathsfit}{bold}{\encodingdefault}{\sfdefault}{bx}{n}

\usepackage{hyperref}
\usepackage{url}
\usepackage{booktabs}
\usepackage{graphicx}
\usepackage{xcolor}
\usepackage{tikz}
\usetikzlibrary{positioning,arrows.meta,fit,backgrounds,decorations.pathreplacing}
\usepackage{fontawesome5} % icons for the overview figure
\usepackage{amsmath}
\usepackage{multirow}
\usepackage{placeins} % \FloatBarrier, keeps appendix floats in their sections

\newcommand{\benchname}{StoreBench}
\title{\benchname{}: A Live-Commerce Environment for Evaluating and
Training Autonomous Operator Agents}

\author{Daksh Raghuvanshi, Ved Vedere, Yifan Wang\corr}
\affiliations{AfterQuery}
\emails{\{daksh,ved,yifan\}@afterquery.com}
\corrnote{Corresponding author: \texttt{yifan@afterquery.com}}
\paperlinks{%
  \plink{Public Set \& Code}{https://github.com/AfterQuery/storebench}%
}

\begin{document}

\maketitle

\begin{abstract}
Reinforcement learning environments are now a primary lever for
improving large language model (LLM) capabilities in post-training, yet most agentic
benchmarks remain static: the world moves only when the agent acts, the
reward is a terminal verdict, and the pass bar is set arbitrarily. We
introduce \benchname{}, a live-commerce environment in which an agent
runs a mid-size online apparel store on a production-grade commerce
backend, testing long-horizon planning and economic judgment under
uncertainty. Customers order around the clock, suppliers reprice and
fail, and market shocks arrive with partial or no warning. The agent
acts through the same 29 merchant tools a human operator would use,
under a windowed operation budget that makes simulated time a function
of actions taken, so model latency cannot influence simulated time. Pass
thresholds are calibrated against scripted anchor policies, the reward
is hardened against a catalogue of reward hacks, and every episode
replays identically given a sequence of actions. We evaluate seven
frontier LLMs on 11 scenarios of 30 to 45 days and a full simulated
year, over three world seeds at matched reasoning effort. No model
matches the scripted smart-triage policy on average: the best,
DeepSeek-V4-Pro, passes 49\% of task-seed cells against the heuristic's
97\%. Human experts working through the same tools and budgets outscore
every model (mean composite 0.708 vs.\ 0.700). Over a full simulated
year under the Claude Code harness, most models show dramatic
performance improvement. In a GRPO post-training run, Qwen3.5-27B
trained on only five disjoint tasks raises its mean composite on the
held-out evaluation tasks from 0.136 to 0.373. We release five example
training-split tasks, ten sample trajectories, and the scoring and
verification tooling; the full environment and evaluation suite are
withheld to keep the benchmark uncontaminated.
\end{abstract}

\section{Introduction}
\label{sec:intro}

Reinforcement learning (RL) with verifiable rewards now drives much of
the progress in reasoning and agentic large language models (LLMs) \citep{deepseekr1,
lambert2024tulu3}. The tasks and evaluators that supply those rewards
therefore increasingly shape what models learn. Meanwhile, AI agents
are moving from answering questions to doing real work: their task
horizon has doubled roughly every seven months \citep{kwa2025horizon},
and evaluations are beginning to cover deliverables drawn from real
occupations \citep{gdpval2025}. Yet most agent benchmarks still test
bounded tasks in controlled environments that are reset between runs
\citep{jimenez2024swebench, zhou2024webarena, yao2024taubench}. As a
result, they rarely measure what ongoing work requires: sustained
reliability, adaptation to change, and the ability to operate with
limited human supervision.

Reliable operations are fundamental to organizational success. Whether
AI agents can excel at them is therefore a critical test of their
real-world capabilities. However, operational work breaks the
conventional evaluation recipe in three ways. First, the world moves
on its own: demand arrives whether or not the agent is ready,
deadlines approach, and information goes stale. The environment's
clock must therefore advance independently of both agent actions and
inference speed. Second, operational success is cumulative, not
binary, and must be measured on a scale calibrated against an external
baseline. Third, open-ended rewards invite open-ended gaming:
mismatches between the simulator's rules and its intended behavior
create loopholes that reward optimization can teach agents to exploit.
Together, these challenges create a need for evaluations of sustained
operations under asynchronous demand, with empirically calibrated
success criteria and rewards explicitly stress-tested for exploits.

\begin{figure}[t]
\centering
\resizebox{\textwidth}{!}{{\aqtimes\input{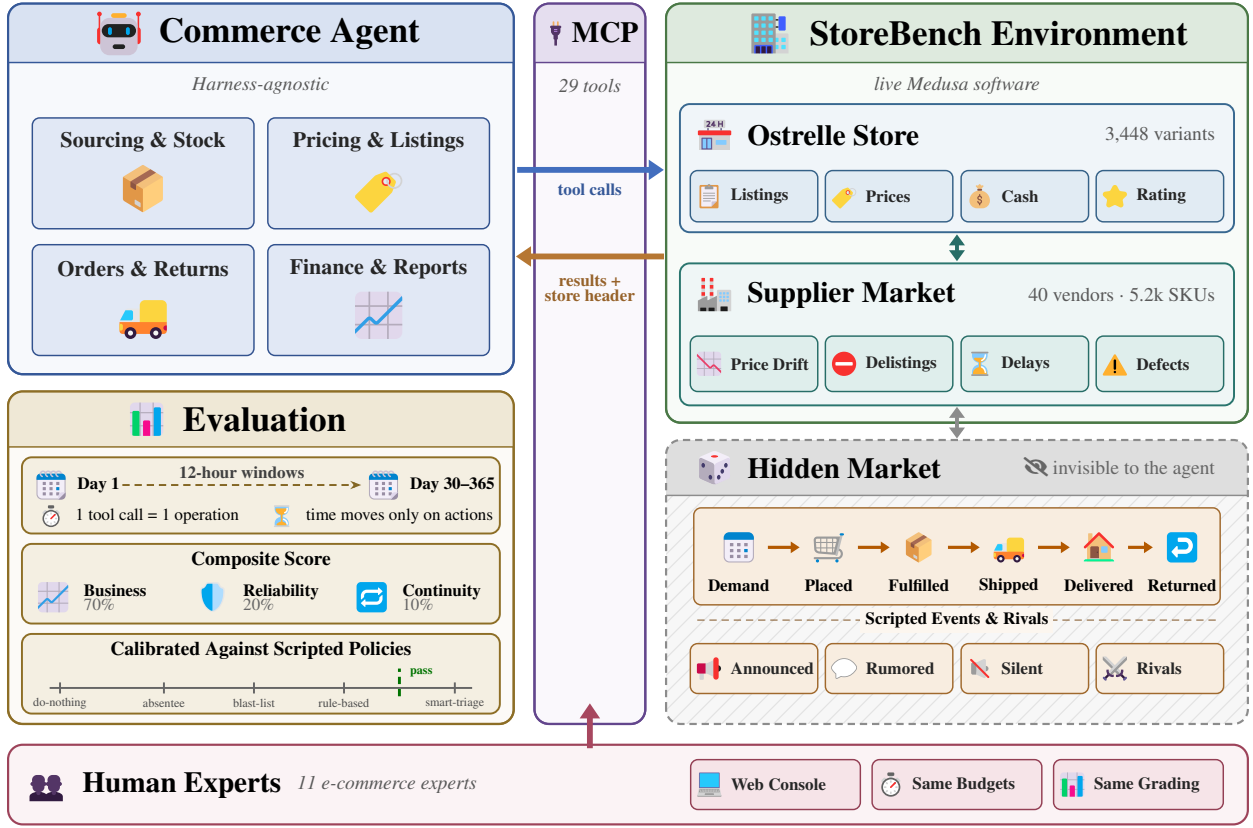}}}
\caption{\benchname{} overview. A harness-agnostic commerce agent
reaches the world only through 29 MCP tools; each call spends one
operation, and the world advances in 12-hour windows. The Ostrelle store
and its supplier market run on live Medusa~v2 software, while demand,
order flow, scripted events, and rivals sit in a hidden market the agent
observes only through its own sales. Episodes of 30 to 365 days receive
weighted scores and are compared to scripted anchor policies.
Human experts operate the same MCP surface through a thin GUI, under
the same budgets and grading. Icons: Microsoft Fluent Emoji (MIT License).}
\label{fig:overview}
\end{figure}

We therefore introduce \textbf{\benchname{}}, a long-horizon benchmark
in which an agent operates a mid-sized online apparel store for 30 to 365
simulated days (Figure~\ref{fig:overview}). The environment combines 18 months of operating
history, a 40-vendor supplier market, 29 merchant tools, hidden demand,
and scripted shocks to suppliers, costs, and competitors within a
production-grade commerce backend. Simulated time depends only on agent
operations, not inference latency. Scripted policies calibrate each
task-seed score and pass threshold, while an exploiter policy tests
resistance to known reward hacking. Our contributions are threefold.
\begin{itemize}
\item \textbf{A realistic agent environment for long-horizon
operational work.} The framework of \benchname{} is general: its
modular tools, event models, and backend can be adapted to other
real-world operational settings.
\item \textbf{Calibrated, exploit-resistant evaluation.} Regression
tests cover every known exploit, any run can be replayed for audit, and
dense rewards make \benchname{} suitable for RL.
\item \textbf{An unsaturated benchmark with actionable failure modes.}
We evaluate frontier models on tasks calibrated to a difficulty that
still separates them: none matches the smart-triage operator on average.
Replayable traces turn that headroom into concrete targets for
improving models and harnesses.
\end{itemize}

\section{Related work}
\label{sec:related}

\paragraph{Agent benchmarks.}
Most agent benchmarks evaluate predefined goals in environments that
change only when the agent acts, including software repair
\citep{jimenez2024swebench}, web and desktop tasks
\citep{deng2023mind2web, zhou2024webarena, xie2024osworld}, tool-use
dialogues \citep{yao2024taubench, vitabench2026}, simulated workplace
tasks \citep{xu2024agentcompany}, and general-assistant suites
\citep{liu2023agentbench, mialon2023gaia}. DAYJOB extends this paradigm
to multi-hour professional assignments graded with expert rubrics
\citep{surge2026dayjob}, but the unit of evaluation remains a fixed
assignment. These benchmarks primarily measure whether an agent
completes the task and reaches an end state, not the quality it sustains as demands change.

\paragraph{Dynamic operational environments.}
Gaia2 introduces scripted events that occur independently of the
agent, but still evaluates discrete assistant tasks against an oracle
action graph \citep{gaia2iclr2026}. Its Time scenarios are capped at
five minutes, and generation latency affects the clock: removing it
raises GPT-5's score from 0\% to 34.4\%. Vending-Bench
\citep{backlund2025vendingbench} and MerchantBench
\citep{merchantbench2026} evaluate long-running businesses using
terminal financial outcomes, without task-specific calibrated scales or
pass thresholds. Together these gaps point to a need for long-horizon
evaluation that meters time through operations rather than generation
latency, runs on a production-grade commerce backend, and calibrates
each task-seed pair against measured policy baselines.

\paragraph{Reward hacking.}
Reward hacking is a longstanding concern in RL safety
\citep{amodei2016concrete, krakovna2020specification,
skalse2022defining, pan2022effects, gao2023scaling}, and LLM agents
have been observed modifying rewards and tests, tampering with evaluators, and exploiting feedback loops
\citep{denison2024sycophancy, pan2024feedbackloops, baker2025monitoring,
metr2025rewardhacking}. Recent benchmarks measure how often such
behavior occurs \citep{zhong2025impossiblebench, zhao2026specbench,
thaman2026rewardhacking}. However, an open-ended reward needs
hardening by design, with each closed exploit fixed by a regression
test and calibration failing whenever an exploiter policy beats an
honest baseline.

\section{The \benchname{} environment}
\label{sec:environment}

\benchname{} places the agent in charge of \emph{Ostrelle}, a fictional
mid-size online apparel store, for a contiguous shift of 30 to 365
simulated days. The store is not a mock: it runs on Medusa~v2, a
production-grade open-source commerce platform \citep{medusa2024}, with
the agent acting through the same administrative API surface a human
operator would use.
One container image holds the backend, simulation engine, and tool
server; the agent's network can reach the tool surface and nothing
else.

\subsection{A world with history}
\label{sec:world}

The store opens with 18 months of coherent operating history baked into
the backend's own database: 3{,}320 orders with seasonal and diurnal
structure, $\sim$2{,}500 customers with realistic repeat-purchase
distributions, a 14.7\% return rate, and opening work-in-progress that
becomes the agent's obligations at window one. A seed-quality suite of
31 distributional and coherence audits gates every published image.

The listed catalog covers 111 supply lines over 3{,}448 variants; a
supplier market of 40 vendors offers $\sim$5{,}200 stock-keeping units (SKUs) across tiers
that differ in price, lead time, minimum quantities, defect rates, and
reliability. Supplier prices drift under a bounded random walk; scripted
events can withdraw SKUs, extend lead times, or reprice whole origins.

\subsection{Hidden dynamics}
\label{sec:dynamics}

Demand is generated per 12-hour window from a hidden model the agent can
only observe through its own sales: per-line popularity, price
elasticity, shelf attractiveness, seasonality, copy quality, and
stockout censoring. Orders are injected into the real
backend and backdated to simulated time, so the fiction survives
direct database inspection.
Returns arrive with realistic delays and escalate to fines if ignored;
defective units follow supplier-specific rates.

Scripted \emph{events} perturb the world on three information channels:
\emph{announced} (a platform alert with lead time), \emph{rumored}
(vague trade-press language, no numbers), and \emph{silent} (only the
store's own sales reveal it). The vocabulary spans demand shifts,
supplier failures, quality crises, macro crashes, cost inflation, and
country-level supply shocks. A competition layer adds scripted rivals
that capture demand share as a function of relative shelf
attractiveness; a store that prices lazily bleeds share even when
nothing is announced.

\subsection{Tasks as lenses}
\label{sec:tasks}

\begin{figure}[t]
\centering
\resizebox{\textwidth}{!}{{\aqtimes\input{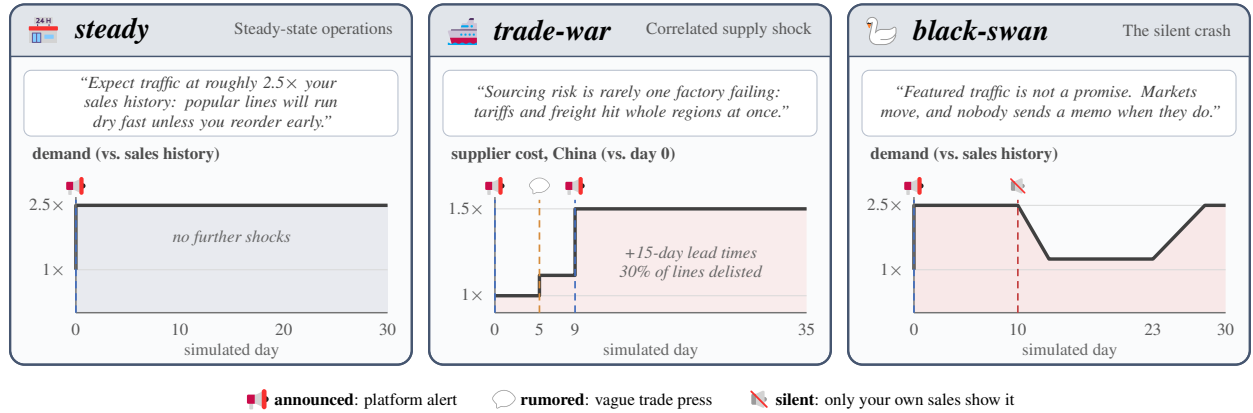}}}
\caption{Three example tasks from the 12-task suite: \textit{steady}
(steady-state operations), \textit{trade-war} (a correlated supply
shock), and \textit{black-swan} (a silent demand crash). Each chart
plots the hidden quantity the task's event script perturbs; markers show
when each event fires and through which channel the agent can learn of
it. Quotes are from the task prompts.}
\label{fig:tasks}
\end{figure}

The benchmark comprises 12 task \emph{lenses} over the same seeded
world; each is a small configuration that sets horizon, operation
budget, and event script (Figure~\ref{fig:tasks};
Appendix~\ref{app:tasks}). They span steady-state competence, cold
starts, sourcing failures, demand regime changes, capital discipline,
competitive defense, compounding shocks, and a 365-day marathon
(\textit{full-year}). All tasks share one world: the catalog, the
hidden market state, and the 18-month history are fixed. Three world
seeds (7, 11, 13) key the forward noise streams; every task-seed pair
is calibrated independently (\S\ref{sec:scoring}), and two further
seeds are held out.

\subsection{The tool surface and metered time}
\label{sec:metering}

The agent interacts exclusively through 29 tools over the Model Context
Protocol, spanning catalog, pricing, inventory, fulfillment, returns,
sourcing, and reporting (Appendix~\ref{app:tools}); every result embeds
a one-line store header with time, window, operations left, cash, net
assets, and rating. Time is metered in \emph{operations}
(Figure~\ref{fig:overview}): each call except a free status probe costs
one, and spending the window's last operation or calling
\texttt{end\_window} advances the world. Budgets do not roll over, so
reading a report and shipping an order compete for the same scarce
operations. Because the world advances only on agent actions, simulated
time is a pure function of the action sequence and wall-clock speed buys
nothing: a slow model under a chatty harness faces the same pressure as
a fast one. Actions execute at the frozen window-start time; writes are
serialized and \emph{fail closed}, so mutations are refused if the
engine is unreachable, and the world freezes at episode end for grading.
One read returns up to 100 rows while one write moves one order, so
observation is comparatively cheap; empirically, final score is flat
across a tenfold range of read volume.

\section{Grader design}
\label{sec:scoring}

\subsection{The composite score}
\label{sec:composite}

An episode is graded on the terminal state of the store's books after a
settlement pass that enforces every obligation that came due within the
horizon (unshipped orders are fined, ignored returns escalate,
in-flight purchase orders realize salvage). The composite is
\begin{equation}
R \;=\; 0.7\,\underbrace{\mathrm{clip}_{[0,1]}\!\left(\frac{\Delta
\mathrm{net}-z}{s}\right)}_{\text{business}}
\;+\;0.2\,\underbrace{\tfrac{1}{2}\left(\text{on-time}+\text{returns
handled}\right)}_{\text{reliability}}
\;+\;0.1\,\underbrace{\frac{\text{windows attended}}{\text{windows in
task}}}_{\text{continuity}},
\label{eq:composite}
\end{equation}
where $\Delta\mathrm{net}$ is the episode's net-asset growth (cash,
deposits, receivables, in-transit money; inventory only at salvage
value), and $z$ and $s$ are the measured zero and scale for the
task-seed pair (\S\ref{sec:anchors}). Reliability counts platform
auto-shipments and stockout cancellations \emph{against} the on-time
rate, and return credit scales with the share of the customer's money
actually refunded.
Continuity divides by the windows in the task, not the windows played:
quitting is absence, not a shorter denominator. Every window also
emits an unclipped dense training channel (\S\ref{sec:posttraining},
Appendix~\ref{app:dense}). The weights encode a judgment about what
matters in store operation; the ranking is stable under alternative
weightings (Appendix~\ref{app:weights}).

\subsection{Anchor calibration}
\label{sec:anchors}

A reward like Eq.~\ref{eq:composite} is meaningless without a measured
scale. For every task-seed pair we run the six scripted policies
through the real tool surface, under the
same budgets as any agent, from \textbf{do-nothing} and
\textbf{absentee} (attends, never acts) through
\textbf{blast-list} (ships everything, decides nothing),
\textbf{rule-based} (a checklist clerk), and \textbf{smart-triage} (a
heuristic operator) to an \textbf{exploiter} that plays every closed
reward exploit.

The anchors define the scale: $z$ is absentee's terminal growth, $s$ the
gap from $z$ to the best anchor. The pass threshold sits at the midpoint
of the two best \emph{honest} trading anchors, subject to four gates.
(i)~\emph{Ordering}: on every seed, both judgment anchors (rule-based,
smart-triage) must outscore the volume and non-trading anchors; their
order relative to each other may vary.
(ii)~A \emph{judgment gap}: the better judgment anchor must clear the
volume policy (blast-list) by a minimum gap, otherwise the lens rewards
keystrokes rather than judgment and is rebalanced.
(iii)~A \emph{flap margin}: a pass bar within run-to-run noise of a
reference policy will flap, the reliability failure documented for
small evaluation sets \citep{microbench2026}, so the threshold must
clear both adjacent anchors by a noise margin (where the top two tie,
the bar sits just under the shared ceiling). (iv)~\emph{Exploit suppression}:
the exploiter must score
below every honest anchor. All 36 reported task-seed cells pass all
gates, as do the 24 cells of the two privately calibrated held-out
seeds; thresholds span 0.43 to 0.76 (Figure~\ref{fig:anchorcells},
Appendix~\ref{app:calibration}). The
scale is demanding: blast-list captures a median 65\% of the top
anchor's composite yet passes nowhere.

\subsection{Reward integrity as engineering}
\label{sec:integrity}

Open-ended rewards invite gaming, and we treat the gap between
``no current model exploits this'' and ``no optimizer can profit from
this'' as the benchmark's attack surface. Development followed an
audit-close-pin loop: each exploit found by an audit of 756
pre-hardening trials and a code-level exploit hunt was closed in the
world model, pinned by an adversarial regression test, and folded into
the exploiter anchor, so every future calibration re-plays the entire
catalogue. Post-hardening, the exploiter averages 0.04 over the reported
suite against 0.46 to 0.76 for the honest trading anchors
(Table~\ref{tab:leaderboard}); the full catalogue, with
measured pre-fix payoffs, is in Appendix~\ref{app:exploits}.

\subsection{Determinism and provenance}
\label{sec:provenance}

Every reported number is reproducible from an immutable chain. Noise
streams are purpose-keyed, replaying an action sequence against the
live backend reproduces the ledger byte for byte, and each trial's run
manifest records the source commit, image digest, and hashes of the
seed data and calibration file. Missing or corrupt artifacts grade to
numeric zero: the environment fails closed at every layer
(Appendix~\ref{app:provenance}).

\section{Experimental setup}
\label{sec:experiments}

\paragraph{Protocol.} We evaluate seven frontier models
(DeepSeek-V4-Pro, Claude~Opus~4.8, Claude~Fable~5, Qwen3.8-Max,
Gemini~3.8~Flash, GPT-5.6 Sol, and Muse~Spark~1.2) on our eval task
suite of 11 dense tasks $\times$ 3 reported seeds (7, 11, 13) $\times$ 3
attempts under a single pinned agent harness (goose) with the 29-tool
MCP surface mounted. The goose harness is open-source and tool-native
by default, which makes it a natural medium for public evaluations.
Every model runs at matched reasoning effort, with full harness and
configuration details in Appendix~\ref{app:config}; five training-split
tasks, ten sample trajectories, and the scoring code are included in the
\textbf{supplementary material}. In every episode, the model
receives the task instruction and tool access, and no scoring
mechanics, thresholds, or scripted-event information. The harness
exposes no web or browser tools, and egress is limited to the model
API. Each individual trial is considered a ``pass'' if its composite
score (Eq.~\ref{eq:composite}) clears the task-seed threshold.

The full-year task is run separately, using Claude Code, a popular
production harness, with context compaction and extended thinking
suited to a 365-window episode, at 3 seeds $\times$ 3 attempts. Because
it differs from the short-horizon suite in both harness and effort, we
report it as its own result (\S\ref{sec:results-horizon}) for fair
comparison. As in the main benchmark, every model faces byte-identical
starting conditions.

\paragraph{Metrics.} We report per-cell composite scores with pass/fail
rates against per-cell calibrated thresholds, component breakdowns
(business, reliability, continuity), and position relative to the
deterministic set of policies. Genuine model failures, like an agent
that quits, idles, or declares itself finished early, are kept and
scored as played; only trials that fail for infrastructure reasons (API
or sandbox errors) are excluded and re-run.

\paragraph{Human expert reference.} Eleven randomly selected
participants from an existing pool of contract experts with reported
e-commerce
experience operated the store on the same 33 cells we report in the
main benchmark (Appendix~\ref{app:human}). They interacted with Ostrelle using a thin client over
the MCP server, to create an identical tool surface and observation
space as their agent counterparts, along with identical operation
budgets, calibration, and grading. Participants received the same
written briefing as the agents' initial observation, and were allotted
12 hours of wall clock time to complete each task. The user interface
constructed for the human experts allowed no privileged information or
advantages, in order to represent tasks with the same conditions that
evaluated agents faced.

\section{Evaluation results}
\label{sec:results}

Under the protocol of \S\ref{sec:experiments},
Table~\ref{tab:leaderboard} reports the leaderboard against the
scripted policy set; the full model~$\times$~task composite matrix is in
Appendix~\ref{app:pertask}.

% AUTO-GENERATED by results_gen.py -- do not edit by hand.
% Regenerated 2026-09 for the canonical matched-effort short-horizon set
% (7 models, 11 tasks x 3 seeds, stratified-random 3/cell; see CANONICAL_RESULTS.md).
\begin{table}[t]
\centering
\caption{StoreBench leaderboard: we report mean composite score, pass rate, and component means (business, reliability, continuity per Eq.~\ref{eq:composite}) over the 11 eval tasks $\times$ 3 seeds $\times$ 3 attempts, at matched reasoning effort; a trial passes if its composite clears its cell's calibrated threshold (\S\ref{sec:anchors}). Deterministic policies are italicized; no model reaches the smart-triage operator on average. Bold marks the best model or human value per column. Dashes: component means are not reported for scripted policies.}
\label{tab:leaderboard}
\small
\setlength{\tabcolsep}{4pt}
\begin{tabular}{lrrrrr}
\toprule
Policy / Model & Composite & Pass & Business & Reliab. & Contin. \\
\midrule
\textit{Smart-triage} & 0.764 & 97\% & -- & -- & -- \\
\textit{Rule-based} & 0.638 & 12\% & -- & -- & -- \\
\textit{Blast-list} & 0.463 & 0\% & -- & -- & -- \\
\textit{Absentee} & 0.100 & 0\% & -- & -- & -- \\
\textit{Do-nothing} & 0.000 & 0\% & -- & -- & -- \\
\textit{Exploiter} & 0.039 & 0\% & -- & -- & -- \\
\midrule
DeepSeek-V4-Pro & 0.700 & \textbf{49\%} & \textbf{0.614} & 0.852 & 1.000 \\
Claude Opus 4.8 & 0.668 & 36\% & 0.573 & 0.834 & 1.000 \\
Claude Fable 5 & 0.629 & 29\% & 0.520 & 0.822 & 1.000 \\
Qwen3.8-Max & 0.587 & 20\% & 0.465 & 0.805 & 1.000 \\
Gemini 3.8 Flash & 0.552 & 15\% & 0.440 & 0.719 & 1.000 \\
GPT-5.6 Sol & 0.538 & 19\% & 0.398 & 0.799 & 1.000 \\
Muse Spark 1.2 & 0.387 & 8\% & 0.226 & 0.644 & 1.000 \\
\midrule
Human experts & \textbf{0.708} & 39\% & 0.599 & \textbf{0.944} & 1.000 \\
\bottomrule
\end{tabular}
\end{table}

\subsection{Main benchmark}
\label{sec:results-main}

The headline is that \emph{no frontier model reaches the smart-triage
operator or human expert on average}. The best model, DeepSeek-V4-Pro,
scores a mean composite of 0.700: well below the smart-triage operator
(0.764), and slightly below the mean human expert score (0.708).
DeepSeek and Claude~Opus~4.8 (0.668) are the only models to clear the
checklist-based policy; Claude~Fable~5 (0.629) sits just under it;
Qwen3.8-Max (0.587), Gemini~3.8~Flash (0.552) and GPT-5.6 Sol (0.538)
land between the checklist and the blast-list volume policy (0.463);
Muse~Spark~1.2 (0.387) falls below blast-list. Pass rates extend the
picture: the smart-triage policy clears the calibrated bar on 97\% of
cells, the best model on 49\%, and every other model on at most 36\%.
The discriminating component is \emph{business}: the net assets the
model accrues over the shift.

\subsection{Human experts}
\label{sec:results-human}

The sampled human outcomes achieve a higher mean composite than any
model we evaluated, 0.708: narrowly above DeepSeek (0.700), well
above the rule-based checklist (0.638), and still below the
smart-triage operator (0.764; Table~\ref{tab:leaderboard}). The gap to
smart-triage may partly reflect the interface: humans operated through
the same tool-and-operation surface as the agents rather than the
storefront tools they normally use. From the observed results, their advantage lies in service rather than commerce.
Humans ship 96\% of orders on time and handle 93\% of returns, for a
reliability of 0.944 against at most 0.852 for any model, while their
business component (0.599) is slightly below DeepSeek's (0.614). Human scores are also more
dispersed across cells (0.33 to 1.00, standard deviation 0.16), so
despite the higher mean they clear the calibrated threshold on fewer
cells than DeepSeek (39\% vs.\ 49\%). By task (Appendix~\ref{app:pertask}),
humans exceed every model on slow-drift, black-swan, new-entrant, and
perfect-storm, exceed even smart-triage on cold-start and slow-drift,
and trail the best model most on steady (0.539 vs.\ 0.736). Two
caveats bound the comparison: humans contribute one selected outcome
per cell where models average three unselected attempts, and
participants were allowed one retry.

\subsection{Long-horizon operation}
\label{sec:results-horizon}

\begin{table}[t]
\centering\small
\caption{Long horizon vs.\ short horizon. $\Delta$ is full-year minus
the eval mean; the setups differ in harness.}
\label{tab:fullyear}
% AUTO-GENERATED by make_horizon_figure.py from gen_fullyear.tex
% -- do not edit by hand.
\begin{tabular}{lrrr}
\toprule
Model & Short & Full-yr & $\Delta$ \\
\midrule
DeepSeek-V4-Pro & 0.700 & 0.976 & +0.276 \\
Qwen3.8-Max & 0.587 & 0.974 & +0.387 \\
Claude Opus 4.8 & 0.668 & 0.971 & +0.303 \\
Claude Fable 5 & 0.629 & 0.966 & +0.337 \\
Muse Spark 1.2 & 0.387 & 0.862 & +0.475 \\
GPT-5.6 Sol & 0.538 & 0.783 & +0.245 \\
Gemini 3.8 Flash & 0.552 & 0.496 & -0.056 \\
\bottomrule
\end{tabular}

\end{table}

We evaluate our seven models on the \emph{full-year} task
(Table~\ref{tab:fullyear}), where models
operate the store for 365 days, one window a day, twelve operations in
the window, and no scripted shock. Orders, returns, and fines continue
to be simulated year-round, with seasonal shifts in demand and carrier
availability. It is run under Claude Code, with context compaction and
extended thinking, rather than the goose protocol of the short suite.

The short-horizon order does not carry over, and most models score
higher. DeepSeek-V4-Pro (0.976), Qwen3.8-Max (0.974), Claude~Opus~4.8
(0.971), and Claude~Fable~5 (0.966) sit near the top of the scale, well
above the smart-triage anchor (0.764). Those four gain +0.28 to +0.39 on
their short-horizon means. Muse~Spark~1.2, last on the short suite at
0.387, reaches 0.862. Qwen, fourth over the short tasks, is second here.
The deltas compare the two setups, harness and effort included. Gemini~3.8~Flash
(0.496) is the one model whose year falls below its short-horizon
mean; GPT-5.6 Sol still rises, to 0.783, short of the leaders. Both
scores include attempts that fail outright
(Figure~\ref{fig:horizonflip}). In four of nine Gemini attempts, the
agent asserts that a background simulation process is running the store
and then waits through the harness's scheduling tool instead of acting,
averaging about 0.1; on the five attempts where it operates, it averages
0.799. One of nine GPT attempts declares the store complete at window
185 of 365 and stops. Both failures are scored as played.

\subsection{Failure modes}
\label{sec:results-failures}

The replayable traces let us understand models' specific failure modes,
using both recorded traces and agents' own transcripts. Across the
official benchmark suite, three failure modes dominate the business
term, and all stem from a misallocation of a scarce operating budget.
\begin{itemize}
\item \textbf{Underbuying and stockouts.} Restocking that is too thin,
too late, or skips secondary lines empties the shelf at the SKU level
and forfeits demand. It appears in 26\% of trials (178 of 693) and spans
the whole field, including the highest-scoring models; in 60 trials the
agent never places a single purchase order (Gemini in 41 of them). One
Gemini \emph{trade-war} run places zero restocks and lets inventory
drain from 1{,}603 units to zero, losing 61 orders; even a strong Opus
run narrates the pattern verbatim: ``second-tier sellers are stocking
out \ldots I never restocked them.''
\item \textbf{Return under-refunding.} Agents do engage returns, but
refund below the full amount: the customer rejects the partial refund,
the return escalates past its deadline, and the store pays an
ignored-return fine, eventually a worse outcome than the refund it was
avoiding.
\item \textbf{Shell-driving vs.\ MCP-use.} Given identical access,
models may issue native tool calls or script batched calls through the
harness shell. The two weakest operators, Muse and Gemini, lean on
shell use indiscriminately, dispatching
bulk fulfill/refund/restock batches that act on stale state and spending
the budget on status polls rather than writes. The cost shows up within
each model: Muse's shell-driven runs average 0.34 composite against
0.43 when it operates natively, and Gemini's two native runs average
0.89 against 0.55 for its shell-driven ones. The split follows the
model rather than the task: the two open-weight models, DeepSeek-V4-Pro
and Qwen3.8-Max, call the tools natively in every trial, while
Claude~Opus~4.8, Claude~Fable~5, GPT-5.6 Sol, and Gemini~3.8~Flash drive
the store through the shell in nearly every trial, and Muse~Spark~1.2
splits evenly.
\end{itemize}

\section{Post-training results}
\label{sec:posttraining}

\benchname{} is designed to serve as a training environment as well as a
benchmark: grading is deterministic and judge-free, simulated time is a
pure function of the action sequence (\S\ref{sec:metering}), and a dense
per-window scoring channel telescopes exactly to the graded terminal
growth (Appendix~\ref{app:dense}). We test whether RL in the
environment improves operation on tasks the policy never
trained on.

\paragraph{Setup.} We post-train Qwen3.5-27B with GRPO
\citep{shao2024deepseekmath} and LoRA \citep[rank 32;][]{hu2022lora} on
five 7- to 15-day training tasks (cash-crunch, counter-punch, final-clearance,
season-flip, short-runway) that are disjoint from the evaluation suite,
over seeds 7, 11, and 13. We take 36 training steps of 5 task-seed
prompts $\times$ 6 rollouts, or 1{,}080 total training episodes. The base
model's composite score consistently receives a 0 for the business
component, and it almost never trades at a profit, so rollouts within a
group scored alike and left no advantage signal. The training reward
therefore clips normalized business to $[-2, 1]$ instead of $[0, 1]$,
which ranks losing stores against each other. Every reported number
still uses the official composite score. We evaluate the base model and
the final checkpoint on the 11 evaluation tasks, whose 30- to 45-day horizons
are at least twice the training length, $\times$ seeds 7, 11, 13 $\times$ 3
episodes. For both training and evaluation, we use a custom in-process
Python harness with a 150-turn cap and context compaction
(Appendix~\ref{app:posttraining}). Because this harness differs from the
leaderboard protocol, we compare the two policies with each other and
with the anchors, not with Table~\ref{tab:leaderboard}.

\begin{table}[t]
\centering\small
\caption{Post-training transfers to held-out tasks. Qwen3.5-27B before
and after GRPO on the 11 evaluation tasks $\times$ 3 seeds $\times$ 3
episodes, official composite score, and its components. On-time
fulfillment rates and successful return rates are also shown. Both
policies run under the same harness.}
\label{tab:posttrain}
\setlength{\tabcolsep}{4pt}
\begin{tabular}{lrrrrrrrr}
\toprule
Policy & Composite & Pass & Business & Reliab. & Contin. & On-time & Returns & Windows \\
\midrule
Base & 0.136 & 0/99 & 0.005 & 0.401 & 0.521 & 0.65 & 0.16 & 33.4 \\
Tuned & \textbf{0.373} & \textbf{3/99} & \textbf{0.236} & \textbf{0.669} & \textbf{0.741} & \textbf{0.89} & \textbf{0.45} & \textbf{47.3} \\
\bottomrule
\end{tabular}
\end{table}

\paragraph{Results.} Post-training raises the mean composite on the
held-out tasks from 0.136 to 0.373, a paired gain of $+0.237$ per cell
(bootstrap 95\% CI $[+0.19, +0.29]$), and improves 31 of 33 cells
(Table~\ref{tab:posttrain}; per-task results in
Appendix~\ref{app:posttraining}). The share of episodes that end at a
profit rises from 3\% to 72\%. The trained policy ships more orders on
time and clears more returns. The base model exhausts the turn cap on 97
of 99 episodes and reaches window 33 on average (tasks run 60 to 90
windows); the checkpoint reaches window 47. Neither policy reprices or
restocks much, and the checkpoint places fewer purchase orders than the
base model. The trained policy still exceeds the blast-list anchor on
only two tasks (liquidity and steady), stays below the rule-based
checklist on all eleven, and passes 3 of 99 episodes. Both regressing
cells are on cold-start, where in 8 of 99 episodes the checkpoint stops
calling tools after about twelve windows, which the base model never
does.

\section{Discussion}
\label{sec:discussion}

\paragraph{Human experts versus models.} Humans reach their higher mean mainly
through service, shipping and refunding more reliably than any model
while earning slightly less than the best one. The remaining gap lies
less in knowing what a good store does than in executing routine
obligations consistently. The failure of models to reach human-expert
play, despite humans interacting with an unnatural interface, shows the
gap the models have yet to bridge with respect to long-term coherence
and execution.

\paragraph{Harness as a variable.} Under Claude Code, five of
seven models score 0.86 or higher on the full year, above both their
short-horizon means and smart-triage (Table~\ref{tab:fullyear}). The
short-horizon suite (60 to 90 windows, goose harness) and the full-year
task (365 windows, Claude Code) differ in harness as well as horizon, so
we report them separately, do not average across them, and cannot
isolate the harness's effect. The size of the gains nonetheless suggests
that the harness is a potential confound, one often left unexplored by
public benchmarks. We propose that future public benchmarks should,
whenever possible, report their scores relative to performance using a
state-of-the-art harness.

\paragraph{Shell versus tool.}\label{sec:discussion-shell} Every model
saw the same 29 tools plus a shell, and a call costs one operation
either way, yet four of the five closed models drive the store almost
entirely through the shell, a preference that grew with reasoning effort
in internal tests. One possible explanation is extensive post-training
on terminal-based tasks; when and why agents prefer this interface
remains open.

\paragraph{Post training gains.} The checkpoint's gains come
primarily from operating discipline rather than commercial strategy
(more on-time shipments and handled returns, little repricing or
restocking) and partly from turn efficiency under the 150-turn cap.
Learning the pricing and sourcing judgment that separates the strongest
models likely requires longer training. That five shorter-horizon
training tasks yield meaningful held-out gains also invites study of
sample efficiency on long-horizon tasks.

\paragraph{Limitations.} The store is a single apparel business with
scripted competitors and parametric demand; the engine is withheld to
keep the suite held back; and post-training uses one base model and
one run (Appendix~\ref{app:limitations}).

\section{Conclusion}
\label{sec:conclusion}

We presented \benchname{}, a live-commerce environment that grades
sustained economic operation of a real store under hidden dynamics, with
harness-independent metered time, anchor-calibrated scoring, an
exploit-hardened reward surface, and a byte-auditable measurement chain.
No frontier model matches the smart-triage anchor on average, and
although most sustain operation well above it over a full simulated
year, the long horizon exposes failure modes that short tasks never do:
silent delegation to a hallucinated background process and premature
self-declared completion.

\subsubsection*{Reproducibility statement}
Five example tasks from the training split, ten sample trajectories on
them, their calibration, and the scoring and verification code (the
composite scorer and a standalone re-grader that reproduces each
trajectory's score from its exported ledger) are included as anonymized
supplementary material, with a datasheet and an MIT license. Because the
benchmark is a held-back test set, we withhold the simulation engine and
the entire evaluation suite (every evaluation task's prompt,
configuration, and calibration), so the environment cannot be run or
trained against outside our own evaluation; the released scorer and
trajectories let others reproduce a graded score from an exported ledger
without the engine. Each released trajectory carries its graded terminal
ledger and a redacted run summary. The determinism and provenance
mechanisms that make every reported number replayable are described in
\S\ref{sec:provenance}; the metering that makes results
harness-independent in \S\ref{sec:metering}; the calibration procedure
and its gates in \S\ref{sec:anchors}, with the reported suite summarized
in Appendix~\ref{app:calibration} and Appendix~\ref{app:tasks}; and the
harness, reasoning-effort, and trial-selection settings in
Appendix~\ref{app:config}.

\subsubsection*{AI use statement}
Large language models are the \emph{subjects} of this benchmark; all
evaluated model outputs appear only as measured results. In this work,
we additionally used generative AI tools for implementing methods
(portions of the environment, calibration, and analysis code were
written with AI coding assistants), for generating synthetic datasets
(the seeded store history was produced by procedural generators
developed with AI assistance and validated by a 31-check audit suite),
for feedback on research methodology (calibration-gate design was
refined in dialogue with AI tools), and for interpreting results (an LLM
classified agent trajectories into the failure modes reported in
\S\ref{sec:results-failures}). We have not used generative AI tools to
formulate mathematical claims or proofs, and translation is not
applicable to this work. Additionally, we used generative AI tools to
draft and edit parts of this manuscript, to create figures, and for
retrieval and discovery (finding and verifying related work and citation
details). We have reviewed all AI-assisted work: environment code is
verified by the unit and adversarial regression suites and by CI gates
on every published image, generated data is validated by the
seed-quality audits of \S\ref{sec:world}, and all claims, failure-mode
counts, and citations were checked by the two lead authors against
exported artifacts and original sources. We take responsibility for the
final content of this work, including text, claims, and artifacts
produced with the aid of generative AI.

\subsubsection*{Ethics statement}
The environment is fully synthetic: no real customers, transactions, or
personal data are involved, and seeded customer records are procedurally
generated. The human expert reference involves paid participants with e-commerce
experience, recruited from an existing contractor pool, who consented to
take part and operated the fictional store; each was fairly compensated for their time. Recorded trajectories contain only actions in the
simulated store and are pseudonymized, with names, contact details, and
access links removed.

\bibliography{references}
\bibliographystyle{afterquery}

\appendix
\appendixcontents
\section{Task suite}
\label{app:tasks}

% AUTO-GENERATED by scripts/make_tables.py - do not edit by hand.
\begin{table}[h]
\centering\small
\caption{The \benchname{} task suite. All tasks share the seeded world; each lens sets horizon, operation budget, and event script.}
\label{tab:tasks}
\begin{tabular}{llrrrrc}
\toprule
Task & Scenario & Days & Windows & Ops/win & Events & Competition\\
\midrule
\textit{steady} & Steady-state operations & 30 & 60 & 20 & 1 & --\\
\textit{cold-start} & Rebuild the shop & 30 & 60 & 12 & 1 & --\\
\textit{supply-shock} & Your biggest supplier fails & 30 & 60 & 20 & 4 & --\\
\textit{quality-crisis} & Find the bad batch & 30 & 60 & 20 & 2 & --\\
\textit{liquidity} & Run it on fumes & 30 & 60 & 11 & 0 & --\\
\textit{peak-season} & The platform festival & 30 & 60 & 16 & 2 & --\\
\textit{slow-drift} & The season turns & 45 & 90 & 12 & 8 & --\\
\textit{black-swan} & The silent crash & 30 & 60 & 20 & 2 & --\\
\textit{new-entrant} & Defend the segment & 30 & 60 & 20 & 1 & yes\\
\textit{trade-war} & Correlated supply shock & 35 & 70 & 20 & 3 & --\\
\textit{perfect-storm} & Everything at once & 35 & 70 & 20 & 3 & yes\\
\textit{full-year} & The marathon & 365 & 365 & 12 & 0 & --\\
\bottomrule
\end{tabular}
\end{table}

Table~\ref{tab:tasks} lists the 12 tasks. Each task's instruction
addresses the agent as the store's operator and
discloses only what the platform would disclose: announced events arrive
with lead time, rumored events as vague trade press, silent events not
at all. Evaluation-task instructions are withheld with the evaluation suite;
the \textbf{supplementary material} includes the prompts and configurations
for the five training-split tasks of \S\ref{sec:posttraining}
(\textit{cash-crunch}, \textit{counter-punch}, \textit{final-clearance},
\textit{season-flip}, \textit{short-runway}), which are a separate set
from the evaluation tasks.

\section{Tool surface}
\label{app:tools}

The 29 tools, by family. Every call costs one operation except
\texttt{store\_status}; \texttt{end\_window} advances the world clock.

\begin{itemize}
\item \textbf{Status and time:} \texttt{store\_status} (free),
\texttt{end\_window}.
\item \textbf{Catalog and listings:} \texttt{list\_products},
\texttt{get\_product}, \texttt{set\_price}, \texttt{set\_listing\_status},
\texttt{update\_listing\_copy}, \texttt{list\_supply\_product}.
\item \textbf{Inventory:} \texttt{list\_inventory}, \texttt{dispose\_stock}.
\item \textbf{Orders:} \texttt{list\_orders}, \texttt{get\_order},
\texttt{fulfill\_order}, \texttt{cancel\_order}.
\item \textbf{Returns:} \texttt{list\_returns}, \texttt{get\_return},
\texttt{refund\_return}.
\item \textbf{Sourcing:} \texttt{search\_supply\_catalog},
\texttt{supply\_product\_detail}, \texttt{supplier\_profile},
\texttt{place\_purchase\_order}, \texttt{cancel\_purchase\_order},
\texttt{list\_purchase\_orders}.
\item \textbf{Finance and market:} \texttt{finance\_summary},
\texttt{ledger\_entries}, \texttt{sales\_report},
\texttt{platform\_terms}, \texttt{announcements},
\texttt{competitor\_watch}.
\end{itemize}

Every tool acts on or reads the live store; there are no decoy tools.
Listing copy feeds a bounded merchandising factor in the demand model,
so the one historically inert tool has measurable effect.

\section{The dense reward channel}
\label{app:dense}

The benchmark grades once, at settlement. For undiscounted
policy-gradient training the choice of reward is free: terminal return
and summed dense reward are the same quantity by construction. The two
diverge under discounting or bootstrapped value learning, where raw
cash flow misprices the environment's central investment: a purchase
order moves cash to in-transit money (neutral), disappears as cost when
it lands, and pays back only as later sales, so a discounted learner
reading cash flow is paid to milk the shelf and starve restocks. The
exported channel therefore values the shelf at salvage,
$\mathrm{rl}_t = \text{net assets}_t + \rho\cdot\text{inventory}_t$
with $\rho$ the salvage rate. Buying stock is neither punished (the
cash outflow is offset when the goods land) nor rewarded (stock is
carried below cost until it actually sells), and because settlement
liquidates leftovers at exactly $\rho$, the channel terminates at the
benchmark number.

The channel is the business term only: reliability and continuity, 0.3
of the composite, are terminal-rate quantities, so training purely on
the dense channel optimizes wealth rather than the reported score. The
clipped business term is flat above the top anchor and below the
absentee anchor by construction; the dense channel is unclipped
precisely so gradient survives outside the calibrated range.

\section{Composite weights: sensitivity}
\label{app:weights}

The composite weights (Eq.~\ref{eq:composite}) encode a judgment about
what matters in store operation; this section checks how much the model
ranking depends on it. Because each cell's composite is linear in its
component scores, mean composites under any weighting recombine exactly
from the exported component means. Table~\ref{tab:weights} reports the
leaderboard under the shipped weights (0.7/0.2/0.1 over
business/reliability/continuity), a reliability-heavy alternative
(0.5/0.4/0.1), and business only.

\begin{table}[h]
\centering\small
\caption{Mean composite under alternative weightings
(business/reliability/continuity). The ranking is essentially stable
across weightings.}
\label{tab:weights}
\begin{tabular}{lrrr}
\toprule
Model & 0.7/0.2/0.1 & 0.5/0.4/0.1 & Business only\\
\midrule
DeepSeek-V4-Pro & 0.700 & 0.748 & 0.614\\
Claude Opus 4.8 & 0.668 & 0.720 & 0.573\\
Claude Fable 5 & 0.629 & 0.689 & 0.520\\
Qwen3.8-Max & 0.587 & 0.655 & 0.465\\
Gemini 3.8 Flash & 0.552 & 0.608 & 0.440\\
GPT-5.6 Sol & 0.538 & 0.619 & 0.398\\
Muse Spark 1.2 & 0.387 & 0.471 & 0.226\\
\bottomrule
\end{tabular}
\end{table}

The top four and the bottom rank are unchanged under all three
weightings; only adjacent, closely-spaced middle models reorder under
the reliability-heavy weighting. Pass rates are not recomputed because
thresholds are calibrated for the shipped composite.

\section{Calibration tables}
\label{app:calibration}

% AUTO-GENERATED by scripts/make_tables.py - do not edit by hand.
\begin{table}[h]
\centering\small
\caption{Anchor composites and calibrated pass thresholds for every task-seed cell (seed 7 / 11 / 13). BL = blast-list, RB = rule-based, ST = smart-triage; the threshold separates the top two honest anchors under the margin and tie gates of \S\ref{sec:anchors}.}
\label{tab:calibration}
\begin{tabular}{lcccc}
\toprule
Task & BL & RB & ST & Threshold\\
\midrule
\textit{steady} & 0.49/0.51/0.45 & 0.67/0.63/0.66 & 0.77/0.77/0.77 & 0.72/0.70/0.72\\
\textit{cold-start} & 0.10/0.10/0.10 & 0.13/0.13/0.13 & 0.74/0.75/0.74 & 0.43/0.44/0.44\\
\textit{supply-shock} & 0.49/0.53/0.45 & 0.69/0.66/0.63 & 0.77/0.77/0.77 & 0.73/0.71/0.70\\
\textit{quality-crisis} & 0.50/0.52/0.43 & 0.71/0.65/0.65 & 0.77/0.77/0.77 & 0.74/0.71/0.71\\
\textit{liquidity} & 0.50/0.58/0.52 & 0.75/0.72/0.76 & 0.75/0.76/0.75 & 0.74/0.74/0.74\\
\textit{peak-season} & 0.45/0.50/0.52 & 0.69/0.69/0.65 & 0.76/0.76/0.77 & 0.73/0.73/0.71\\
\textit{slow-drift} & 0.56/0.47/0.52 & 0.73/0.74/0.70 & 0.77/0.77/0.77 & 0.75/0.76/0.73\\
\textit{black-swan} & 0.50/0.62/0.45 & 0.75/0.77/0.69 & 0.77/0.77/0.77 & 0.76/0.76/0.73\\
\textit{new-entrant} & 0.48/0.53/0.53 & 0.73/0.65/0.64 & 0.77/0.77/0.77 & 0.75/0.71/0.71\\
\textit{trade-war} & 0.46/0.46/0.43 & 0.64/0.61/0.63 & 0.78/0.77/0.77 & 0.71/0.69/0.70\\
\textit{perfect-storm} & 0.52/0.50/0.49 & 0.77/0.67/0.73 & 0.70/0.77/0.77 & 0.74/0.72/0.75\\
\textit{full-year} & 0.41/0.44/0.40 & 0.77/0.77/0.68 & 0.76/0.76/0.77 & 0.75/0.75/0.73\\
\bottomrule
\end{tabular}
\end{table}

\begin{figure}[htbp]
\centering
\includegraphics[width=0.78\textwidth]{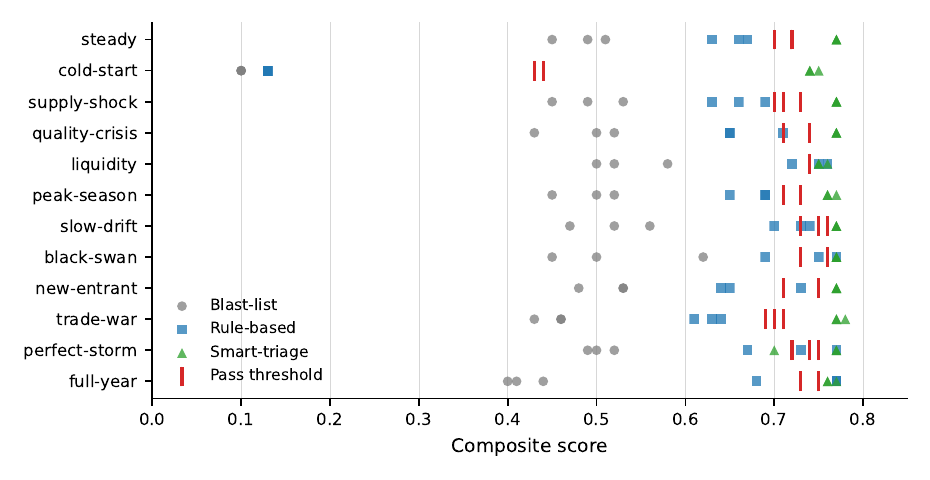}
\caption{The anchor ladder, measured per task-seed cell (one marker per
seed). The calibrated pass threshold sits between the two best honest
anchors on every cell; \textit{cold-start} shows the ladder at its
widest, with checklist play barely above volume play and the threshold
set by the tie-gate against the rung below.}
\label{fig:anchorcells}
\end{figure}

Do-nothing and absentee anchor every cell at composite 0.0 and 0.1
respectively (absentee earns only the continuity term), and the
exploiter is required by gate (iv) of \S\ref{sec:anchors} to land below
every honest anchor; all three are omitted from
Table~\ref{tab:calibration} for space. The gates are enforced, not
aspirational: during development, one candidate lens
(\textit{skeleton-crew}) failed the judgment-gap gate when mass
shipping out-earned triage under its original operation budget, and the
task was rebalanced and recalibrated rather than the gate relaxed.
Figure~\ref{fig:anchorcells} plots the measured ladder for every cell.

\FloatBarrier

\section{The reward-exploit catalogue}
\label{app:exploits}

The catalogue came from two sources: an audit pass over 756
pre-hardening trials and a code-level exploit hunt. It spans classic
specification gaming (free cancellation of open obligations at episode
end, continuity credit for quitting) and subtler economic seams (a
refund-acceptance cliff that made 80-cent-on-the-dollar refunds
optimal). Each entry lists the exploit, its measured pre-fix payoff,
the closure, and the pinning regression test. Every exploit has exactly
one regression test; all tests run in CI on every commit, and the
exploiter anchor re-plays the catalogue at every calibration.

The agent-visible surface was audited for leaks separately: task
identity, world seed, counterfactual demand, and event parameters are
stripped from every tool result, and prompts address a merchant
operating a store, never a model being graded.

\begin{enumerate}
\item \textbf{Continuity for quitters.} Continuity divided by windows
\emph{played}; quitting after 5 of 60 windows scored 1.0 (97\% of
pre-fix trials hit 1.0, including 16 walk-offs). Closed: denominator is
task windows; unplayed windows are absence.
\item \textbf{Reliability laundering.} Platform auto-shipments (fired
exactly because the merchant missed the SLA) were excluded from the
on-time rate; one manual on-time shipment could carry the score.
Closed: auto-shipments and stockout cancels count against on-time.
\item \textbf{Free cancels at the buzzer.} Ending the episode canceled
every open order without fines and dropped them from reliability
(478 pre-fix trials shed $\geq$5 obligations free). Closed: the
settlement pass enforces every obligation due within the horizon.
\item \textbf{Listing side door.} Republishing a delisted product
skipped the insertion fee and the listing cap (201 free publishes across
38 trials). Closed: publishing enforces both, whichever tool is used.
\item \textbf{Lottery pricing.} Shelf demand read the median of all
variant prices, including out-of-stock decoys: plant a cheap phantom
size, sell the real sizes dear. Closed: shelf price reads in-stock
variants only.
\item \textbf{The refund cliff.} Any refund $\geq$ the platform-minimum
share resolved a return with full reliability credit, making
80-cents-on-the-dollar strictly optimal. Closed: credit scales with the
refunded share.
\item \textbf{Phantom cash.} Refunding an unsettled sale created
spendable cash against a receivable that never existed. Closed:
pre-settlement refunds shrink the pending settlement.
\item \textbf{Split-refund trap} (fairness bug, not exploit): the
acceptance floor counted only single refunds, so installments that made
the customer whole never resolved the return (4{,}965 returns across 496
pre-fix trials). Closed: refunds accumulate toward the floor.
\item \textbf{Sub-cent capacity} (fairness bug): payments carry sub-cent
amounts; the default refund rounded up and was refused by the payment
provider, making honest full refunds impossible while lowballs
succeeded. Closed: refund capacity floors to the cent, as the provider
validates it.
\end{enumerate}

\section{Determinism and provenance}
\label{app:provenance}

Noise streams are purpose-keyed, so an unrelated extra action cannot
shift later draws. Replaying an action sequence against the live
backend reproduces the ledger byte for byte, and a CI gate runs this
dual replay on every published image.

Trials record the provenance chain in a run manifest (source commit,
image digest, hashes of the baked seed data and calibration file). Two
links in this chain are enforced, not just recorded: the engine refuses
to arm an episode whose seed data no longer hashes to what CI measured
at build, and the verifier fails any trial whose calibration hash does
not match the score it grades. Trials export the full observation
stream, per-window balances and raw reward deltas, the event log, and
the graded score. Missing or corrupt artifacts grade to numeric zero.

\section{Evaluation configuration}
\label{app:config}

\paragraph{Execution.} Each trial runs in a fresh sandbox from the
digest-pinned image. Network egress is limited to the model API, and the
calibration file and grading endpoints are unreachable in-episode.

\paragraph{Harnesses.} The short-horizon suite runs under goose with
each model's provider context window. Models with a thinking tier
(Claude~Opus~4.8, Claude~Fable~5, GPT-5.6 Sol, Gemini~3.8~Flash) use
goose's highest setting; DeepSeek-V4-Pro, Qwen3.8-Max, and
Muse~Spark~1.2, which reason by default, use their provider default.
goose never compacts context within an episode. The full-year task runs
under Claude Code with maximum extended thinking (32{,}000 tokens per
turn) and context compaction. Muse~Spark~1.2 is reached through a
request-translation proxy, which affects routing only.

\paragraph{Trial selection.} Infrastructure failures are re-run; genuine
model failures are kept. For each model and task-seed cell, three valid
attempts are drawn at random with a fixed seed.

\paragraph{Cost.} Each trial logs its token usage and metered API cost.
Input tokens dominate: the agent re-reads its growing context every
turn, so a short-horizon rollout consumes millions to tens of millions
of input tokens, mostly served from the prompt cache. At first-party
prices, a short-horizon rollout costs about \$1 for GPT-5.6 Sol and
Gemini~3.8~Flash, \$7.5 for Claude~Opus~4.8, and \$31 for
Claude~Fable~5. A full-year rollout costs \$185 on average, from about
\$40 (GPT-5.6 Sol, Muse~Spark~1.2) to \$670 (Claude~Fable~5). The
reported evaluation, about 750 rollouts, cost roughly \$16k in model API
spend, mostly for the full-year task and the Claude models. This is a
lower bound: most short-horizon rollouts for DeepSeek-V4-Pro,
Qwen3.8-Max, and Muse~Spark~1.2 were served through routes that did not
report prices, and sandbox compute is excluded.

\section{Human expert study}
\label{app:human}

\paragraph{Participants and assignment.} Participants were randomly
sampled then recruited from an existing pool of contract experts with
reported prior e-commerce experience; qualifications were self-reported, not
independently verified. Eleven people contributed the 33 reported
outcomes; fifteen have at least one recorded shift. Cells were assigned
adaptively as participants became available, and later assignments
favored participants who had performed well, so contributions are
uneven (one participant supplied ten cells, four supplied one each). The
equal-person mean is 0.724; we report the equal-cell mean (0.708)
throughout.

\paragraph{Interface and support.} The browser console exposed the
agents' tool operations as pages and forms over the same metered
backend, with the same windowed budgets, write freeze, settlement, and
calibration file. It added a history view (a proxy for LLM context), a
returns inbox (to interact with them cleanly), and cached views of
already-fetched data that could be reread without spending an operation
(again, a proxy for context). Before their first shift, participants read a written
briefing on operations, deadlines, sourcing, cash, pricing, and
settlement, which included starting heuristics similar to the
smart-triage anchor's; participants after the initial pilot phase also
completed a short practice task to gain familiarity with the interface. Post-pilot shifts had a 12-hour wall-clock
deadline, and their median span from start to completion was 5.1
hours, including breaks. Participants were fairly compensated
for each accepted shift.

\paragraph{Selection.} Thirty-six shifts reached a final score, and 33
were retained, one per cell. Three cells had an earlier completed
attempt that was replaced: two by coached retries from the same
participant, and one by a different participant after a run affected by
a platform fault. Scoring each cell by its earliest completed attempt
gives a mean of 0.656 instead of 0.708. This difference combines
coaching, reassignment, and the affected run, and is not an estimate of
a training effect. Environment fixes to refund and shipping handling
were released during collection, so the retained shifts span four image
releases; all 33 were graded against the same calibration file.

\section{Per-task and full-year results}
\label{app:pertask}

\begin{figure}[h]
\centering
\includegraphics[width=0.6\textwidth]{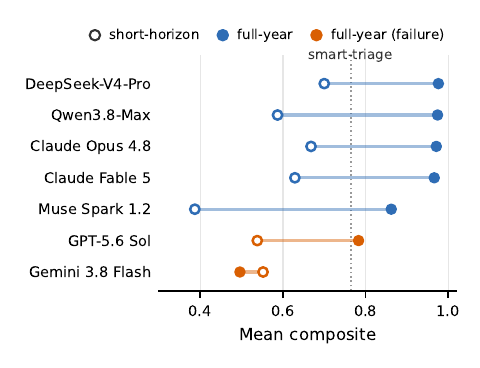}
\caption{Short-horizon mean (open marker, 11 tasks) versus the
365-window \emph{full-year} task (filled marker). Most models operate
well above smart-triage over the full year; the two with observed
long-horizon failure modes (Gemini, GPT) are in orange. Dotted line:
smart-triage.}
\label{fig:horizonflip}
\end{figure}

Table~\ref{tab:permodel} breaks the leaderboard of
Table~\ref{tab:leaderboard} down by task, and
Figure~\ref{fig:horizonflip} plots the short-horizon and full-year
means of Table~\ref{tab:fullyear} side by side.

% AUTO-GENERATED by results_gen.py -- do not edit by hand.
% Regenerated 2026-09: canonical matched-effort short-horizon (11 tasks) + full-year (CC harness).
\begin{table}[htbp]
\centering\small
\caption{Mean composite scores by model and task (3 seeds $\times$ 3 attempts per cell). \emph{Human} is the human expert's score; \emph{ST} is the smart-triage heuristic anchor for reference. \emph{full-year} is measured under a separate agentic harness with extended thinking and is reported separately below eval scores.}
\label{tab:permodel}
\setlength{\tabcolsep}{3pt}
\begin{tabular}{lrrrrrrrrr}
\toprule
Task & DeepS & Opus & Fable & Qwen & Gemini & GPT & Muse & Human & ST \\
\midrule
steady & 0.736 & 0.583 & 0.603 & 0.585 & 0.533 & 0.511 & 0.290 & 0.539 & 0.770 \\
cold-start & 0.851 & 0.632 & 0.855 & 0.767 & 0.545 & 0.466 & 0.692 & 0.794 & 0.741 \\
supply-shock & 0.638 & 0.673 & 0.493 & 0.459 & 0.449 & 0.555 & 0.295 & 0.626 & 0.770 \\
quality-crisis & 0.696 & 0.665 & 0.680 & 0.594 & 0.440 & 0.432 & 0.256 & 0.614 & 0.772 \\
liquidity & 0.723 & 0.689 & 0.670 & 0.754 & 0.513 & 0.645 & 0.552 & 0.733 & 0.753 \\
peak-season & 0.748 & 0.710 & 0.783 & 0.649 & 0.659 & 0.689 & 0.501 & 0.760 & 0.765 \\
slow-drift & 0.741 & 0.687 & 0.652 & 0.563 & 0.631 & 0.620 & 0.357 & 0.799 & 0.768 \\
black-swan & 0.595 & 0.646 & 0.455 & 0.466 & 0.588 & 0.562 & 0.405 & 0.753 & 0.768 \\
new-entrant & 0.692 & 0.600 & 0.581 & 0.458 & 0.564 & 0.459 & 0.330 & 0.717 & 0.769 \\
trade-war & 0.635 & 0.717 & 0.614 & 0.633 & 0.553 & 0.480 & 0.224 & 0.706 & 0.774 \\
perfect-storm & 0.651 & 0.742 & 0.545 & 0.525 & 0.596 & 0.501 & 0.344 & 0.746 & 0.748 \\
\midrule
full-year & 0.976 & 0.971 & 0.966 & 0.974 & 0.496 & 0.783 & 0.862 & -- & 0.764 \\
\bottomrule
\end{tabular}
\end{table}

\FloatBarrier

\section{Post-training details}
\label{app:posttraining}

\paragraph{Training.} GRPO with group size 6 and 5 prompts per step,
advantages normalized by the group standard deviation, a clipped
token-mean policy loss, and no KL penalty to the reference model. LoRA
rank and $\alpha$ are 32. The schedule was adjusted by hand during the
run: the official reward drove steps 1--4, which produced no learning
signal, and the shaped reward of \S\ref{sec:posttraining} drove steps
5--36; the learning rate was $5\times10^{-5}$, raised to $10^{-4}$ for
steps 16--26; steps 5--20 took ten gradient updates per batch and steps
21--36 one. Rollouts used temperature 1.0 and the same harness as
evaluation, with the policy writing its own context summaries. The
official reward on training batches rose from 0.22 over steps 1--4 to
0.61 over steps 27--36.

\paragraph{Evaluation.} Table~\ref{tab:posttrain-tasks} gives per-task
results. Each cell reports its first three valid
episodes; episodes cut by infrastructure faults (API failures,
timeouts, sandbox errors) were rerun, never scored as partial. The
harness caps an episode at 150 turns and compacts context at about 65k
tokens, keeping the last three turns verbatim. Cell means average into
task and suite means, and the confidence interval is a bootstrap over
the 33 paired cell means. Checkpoint 34, evaluated under the same
protocol, scores 0.341 and also improves 31 of 33 cells, so the result
does not hinge on the final checkpoint.

\paragraph{Side effects.} Policy entropy on training batches rose from
0.2 to between 1.0 and 1.9, and the checkpoint's messages drifted
toward an emphatic, occasionally garbled style (19\% of turns vs.\ under 1\%
for the base model). Together with the idle stops of
\S\ref{sec:posttraining}, this suggests that a KL penalty or
lower-temperature sampling should accompany longer runs.

\paragraph{Search baselines.} Before policy optimization, we checked
that cheap search cannot game the composite: on \emph{steady} (seed 7),
300 random-search and 300 hill-climbing episodes topped out at 0.21 and
0.40 against a 0.72 threshold.

\begin{table}[h]
\centering\small
\caption{Post-training by held-out task: mean official composite over 3
seeds $\times$ 3 episodes for the base model and GRPO checkpoints 34
and 36, with the blast-list (BL) and rule-based (RB) anchors on the same
cells.}
\label{tab:posttrain-tasks}
\begin{tabular}{lrrrrrr}
\toprule
Task & Base & Step 34 & Step 36 & $\Delta_{36}$ & BL & RB \\
\midrule
steady & 0.129 & 0.353 & 0.574 & $+0.445$ & 0.481 & 0.656 \\
cold-start & 0.110 & 0.089 & 0.082 & $-0.028$ & 0.100 & 0.133 \\
supply-shock & 0.134 & 0.289 & 0.353 & $+0.220$ & 0.488 & 0.659 \\
quality-crisis & 0.127 & 0.435 & 0.382 & $+0.255$ & 0.483 & 0.669 \\
liquidity & 0.205 & 0.573 & 0.653 & $+0.449$ & 0.536 & 0.741 \\
peak-season & 0.155 & 0.443 & 0.409 & $+0.254$ & 0.492 & 0.674 \\
slow-drift & 0.128 & 0.172 & 0.220 & $+0.092$ & 0.519 & 0.724 \\
black-swan & 0.142 & 0.306 & 0.291 & $+0.148$ & 0.524 & 0.735 \\
new-entrant & 0.135 & 0.435 & 0.448 & $+0.313$ & 0.516 & 0.676 \\
trade-war & 0.114 & 0.292 & 0.361 & $+0.247$ & 0.450 & 0.626 \\
perfect-storm & 0.116 & 0.365 & 0.330 & $+0.213$ & 0.505 & 0.724 \\
\midrule
Mean & 0.136 & 0.341 & 0.373 & $+0.237$ & 0.463 & 0.638 \\
\bottomrule
\end{tabular}
\end{table}

\section{Limitations}
\label{app:limitations}

\paragraph{Scope.} The claim is a faithful direct-to-consumer apparel
store, not a full merchant economy, and the boundaries are deliberate:
demand moves only through the shelf (no advertising channel, no demand
momentum, no market price discovery), competitors are scripted rather
than learned, and the hidden demand model, while elastic, is ultimately
parametric. The full environment is not released: we share a few
training-split tasks, example trajectories, and the scoring code, and
withhold the simulation engine and evaluation suite so the benchmark
stays a held-back test set. The composite's weights
(Eq.~\ref{eq:composite}) encode a judgment; the ranking is stable under
alternative weightings (Appendix~\ref{app:weights}). Scripted anchors
bound checklist and heuristic play, not optimal play; the headroom above
smart-triage is a floor on what stronger operators could show.

\paragraph{Post-training study.} The post-training study uses one base
model, one hand-adjusted training run, and held-out tasks rather than
held-out seeds; it shows that the environment supports hill-climbing
with transfer to unseen scenarios, not a tuned training recipe.

\end{document}